\documentclass{article}
\usepackage{ijcai19}

\usepackage{times}
\usepackage{soul}
\usepackage{url}
\usepackage[hidelinks]{hyperref}
\usepackage[utf8]{inputenc}

\usepackage{multirow}
\usepackage{enumerate}
\usepackage{subcaption}

\usepackage[ruled,noend,linesnumbered,commentsnumbered]{algorithm2e}
\usepackage{amsmath,amsfonts}
\usepackage{amssymb}

\usepackage{wrapfig}
\usepackage{tikz}

\usepackage{tikz-qtree}
\usepackage{tikz,pgf}
\usetikzlibrary{arrows}
\usetikzlibrary{calc}
\usetikzlibrary{positioning}
\usetikzlibrary{arrows, shadows, backgrounds, calc, shapes, decorations, patterns, decorations.pathmorphing, snakes}
\tikzstyle{arg}=[draw, thick, circle]

\newif\ifdraft\drafttrue 
\newif\ifinlineref\inlinereffalse
\newif\iffinal\finalfalse
\newif\ifextended\extendedtrue \extendedfalse 
\newif\ifdotikz\dotikzfalse

\usepackage{macros}

\newcommand{\Lits}{\mathcal{A}}

\newcommand{\rel}[0]{\mi{rel}}
\newcommand{\absrel}[0]{\widehat{\mi{re}}\mi{l}}

\newcommand{\myshift}[0]{\mi{sh}}

\newcommand{\reltype}[1]{\ensuremath{\tau_{\mathrm{#1}}}}
\newcommand{\Reltype}[0]{\ensuremath{\mathcal{T}}}

\finaltrue
\draftfalse

\iffinal

\usepackage[disable]{todonotes}
\else
\fi
\ifdraft
\usepackage{todonotes}
\newcommand{\comment}[1]{{\bf\color{blue}{*** #1 ***}}}
\newcommand{\draftred}[1]{{\color{red}{#1}}}
\else
\usepackage[disable]{todonotes}
\newcommand{\comment}[1]{}
\newcommand{\draftred}[1]{}
\fi

\def\eqs{\,{=}\,}

\def\lars{\,{\lar}\,}
\def\ins{\,{\in}\,}
\def\notins{\,{\notin}\,}
\def\cups{\,{\cup}\,}
\def\mids{\,{\mid}\,}

\def\setminuss{\ {\setminus}\,}

\newtheorem{thm}{Theorem}
\newtheorem{prop}[thm]{Proposition}

\newtheorem{defn}{Definition}

\newcommand{\leanparagraph}[1]{\smallskip\noindent\textbf{#1}. } 
\newcommand{\emleanparagraph}[1]{\smallskip\noindent{\em#1}. } 

\newcommand{\nop}[1]{}

\usepackage{graphicx}

\title{Abstraction for Zooming-In to 
Unsolvability Reasons of Grid-Cell Problems}
\author{Thomas Eiter, Zeynep G. Saribatur, Peter Sch{\"{u}}ller
    \affiliations
    Institute of Logic and Computation, TU Wien 
}

\begin{document}
\maketitle
\begin{abstract}
Humans are capable of abstracting away irrelevant details when studying problems. This is especially noticeable for problems over grid-cells, as humans are able to disregard certain parts of the grid and focus on the key elements important for the problem.
Recently, the notion of abstraction has been introduced for Answer Set Programming (ASP),
a knowledge representation and reasoning paradigm widely used in problem solving
, with the potential to understand the key elements of a program that play a role in finding a solution.
The present paper takes this further and empowers abstraction
to deal with structural aspects, and in particular with hierarchical
abstraction over the domain.
We focus on obtaining the reasons for unsolvability of problems on grids, and show the possibility to automatically achieve human-like abstractions that distinguish only the relevant part of the grid. A user study on abstract explanations confirms 
the similarity of the focus points in machine vs. human explanations
and reaffirms the challenge of employing abstraction to obtain machine explanations.
\end{abstract}

\section{Introduction}

Abstraction is about focusing on the relevant details and disregarding the irrelevant ones that are
not really needed to be taken into account. Human reasoning and constructing explanations involve the use of abstraction, by reasoning over the models of the world that are built mentally \cite{craik1952nature,johnson1983mental}. As abstraction is a common tool for humans when solving a problem, employing this notion in the way machines solve problems has been repeatedly investigated \cite{sacerdoti1974planning,knoblock1990learning,giunchiglia1992theory,clarke03,saitta2013abstraction}. 
Humans are especially capable of using abstraction to point out the details that cause a problem to be not solvable and provide explanations.
In graph coloring, for instance, if a given graph is non-colorable, finding some subgraph (e.g., a clique) of it which causes the
unsolvability, and not caring about other nodes, 
is a typical abstraction 
a human would do. Empowering the machine with an abstraction capability to obtain human-like machine explanations is one of the challenges of explainable AI.

Answer Set Programming (ASP) \cite{
aspglance11} is a knowledge representation and reasoning paradigm widely
used in problem solving 
thanks to its expressive power and the availability of efficient
solvers \cite{DBLP:journals/aim/ErdemGL16}. ASP has been applied in many areas of AI such as planning, diagnosis and commonsense reasoning.
The expressivity and representation power makes ASP a convenient tool for investigating ways of
applying human-inspired problem solving methods.
Ongoing studies in understanding how ASP programs find a solution (or none) to a problem mainly focus on
debugging answer sets
\cite{brain2007debugging,gebser2008meta,oetsch2010catching} or finding
justifications
\cite{pontelli2009justifications,schulz2013aba,cabalar2014causal}.
Recently, the notion of abstraction was introduced for ASP \cite{zgs19jelia}, by means of clustering the elements of the domain
and automatically constructing an over-approximation of a given
program. The introduced abstraction-\&-refinement methodology
(inspired from CEGAR \cite{clarke03}) starts with an initial
abstraction and refines it repeatedly using hints that are obtained from checking the abstract answer sets, until a concrete solution (or unsatisfiability) is encountered. Employing such an abstraction showed potential for
aiding program analysis as it allows for problem solving over abstract notions, by achieving concrete abstract answer sets
that reflect relevant details only.%
For example, for 
graph coloring
this approach enables
the abstraction described above. 

\usetikzlibrary{patterns}
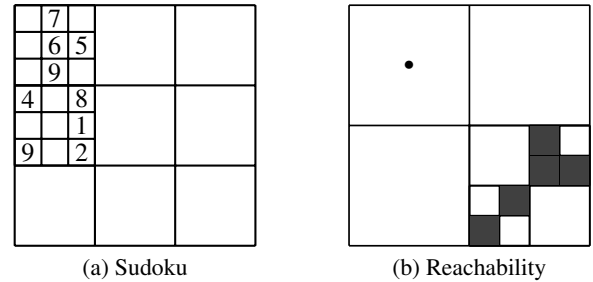
\begin{figure}[t!]
\caption{Unsatisfiable abstractions of grid-cell problems}
\label{fig:unsatgrids}
\centering%
\begin{subfigure}[t]{0.25\textwidth}
\centering
\resizebox{!}{0.14\textheight}{%
\begin{tikzpicture}[scale=0.36]
  \begin{scope}
    \useasboundingbox (0,0) rectangle (9,9);
    \draw[step=3cm,thick, scale=1] (0, 0) grid (9, 9);
    \draw[step=1cm,thick, scale=1,yshift=3cm,xshift=0cm] (0, 0) grid (3, 3);
    \draw[step=1cm,thick, scale=1,yshift=6cm,xshift=0cm] (0, 0) grid (3, 3);
\node[minimum size=0.5cm] at (1.5,8.5) {7};
\node[minimum size=0.5cm] at (1.5,7.5) {6};
\node[minimum size=0.5cm] at (2.5,7.5) {5};
\node[minimum size=0.5cm] at (1.5,6.5) {9};
\node[minimum size=0.5cm] at (0.5,5.5) {4};
\node[minimum size=0.5cm] at (2.5,5.5) {8};
\node[minimum size=0.5cm] at (0.5,3.5) {9};
\node[minimum size=0.5cm] at (2.5,3.5) {2};
\node[minimum size=0.5cm] at (2.5,4.5) {1};

  \end{scope}
    
\end{tikzpicture}}%
\caption{Sudoku}
\label{fig1-b}
\end{subfigure}%
\begin{subfigure}[t]{0.25\textwidth}
\centering
\resizebox{!}{0.14\textheight}{%
\begin{tikzpicture}[scale=0.5]

  \begin{scope}
    \useasboundingbox (0,0) rectangle (8,8);
    \draw[step=4cm,thick, scale=1] (0, 0) grid (8, 8);
    \draw[step=1cm,thick, scale=1,yshift=2cm,xshift=6cm] (0, 0) grid (2, 2);
    \draw[step=2cm,thick, scale=1,xshift=4cm] (0, 0) grid (4, 4);
    \draw[step=1cm,thick, scale=1,yshift=0cm,xshift=4cm] (0, 0) grid (2, 2);
\node at (2,6) {$\bullet$};
\node[fill=darkgray,minimum size=0.47cm] at (4.5,0.5) {};
\node[fill=darkgray,minimum size=0.47cm] at (5.5,1.5) {};
\node[fill=darkgray,minimum size=0.47cm] at (6.5,2.5) {};
\node[fill=darkgray,minimum size=0.47cm] at (7.5,2.5) {};
\node[fill=darkgray,minimum size=0.47cm] at (6.5,3.5) {};
  \end{scope}
    
\end{tikzpicture}}
\caption{Reachability}
\label{fig1-c}
\end{subfigure}
\nbls
\nbls
\end{figure}

Problems 
that involve multi-dimensional structures,
e.g.\ grid-cells, require a differentiated view of an abstraction in order to provide insight that is similar to humans, by focusing on certain areas and abstracting away 
the rest.
Sudoku is a well-known problem, where the empty cells need to be filled with the guidance of the given numbers by respecting some constraints. 
Fig.~\ref{fig1-b} shows an instance with the focus on the sub-regions that contain the reason why a solution can not be found. Since the numbers 6,7 appear in the middle column, they can only be assigned to the below region's left column, which is not possible as only one empty cell exists. As another example, Fig.~\ref{fig1-c} shows an instance for the Reachability problem where some cells are not reachable from the upper-left corner due to the obstacles in the focused area. 

In this paper, we empower the approach in \cite{zgs19jelia} to handle such a hierarchical view of 
abstraction that automatically adjusts the granularity towards the relevant details for the problem.
The method is used for \emph{zooming in} to the area that is sufficient for the machine to realize the unsolvability of a problem instance. Distinguishing this area becomes the machine's way of explaining unsolvability, which is then compared with how humans provide explanations. 


Our contributions are briefly summarized as follows:

\begin{myitemize}
\item We introduce
multi-dimensional abstraction mappings over a domain. For this, we ought to modify the previous abstraction method \cite{zgs19jelia} by having an
existential abstraction over the relations, 
in order to enable dealing with elements of different abstraction
layers.
 \item  We extend the abstraction-\&-refinement methodology with handling the structural aspects of grid-cells by using a quad-tree abstraction and consider more sophisticated
 decision making approaches on the refinement to observe its effects on the resulting abstractions.
  \item We use this approach in detecting the
  unsolvability of benchmarks problems involving grid-cells. A user study is conducted to compare the resulting abstractions with human explanations, which showed that such a hierarchic abstraction can
  provide intuitive and ``to the point'' explanations of
  unsolvability. The user study on human explanations also revealed the implicit abstraction capabilities of humans and the acknowledged need for studying the meaning of explanation.
\end{myitemize}

\section{Background}

\leanparagraph{ASP} We adopt a function-free first order language, 
in which
logic programs are finite sets $\Pi$ of rules 
$r$ of the form 

\smallskip

\centerline{$\alpha \lars B(r)$,}

\smallskip

\noindent where $\alpha$ is an atom and 
the body $B(r)= l_1,\ldots,l_n$ is a set 
of positive and negative literals $l_i$ of the form $\beta$ or
$\mi{not}\ \beta$, resp., where $\beta$ is an atom and $\mi{not}$ is default negation;
$B^+(r)$ and $B^-(r)$ are the sets of all positive resp.\ negative
literals in $B(r)$. A rule 
$r$ is a \emph{constraint},
if $\alpha$ is falsity ($\bot$, then omitted). 
\emph{Choice rules} of the form
$\{\alpha\} \lars B$ are a shorthand for 
$\alpha \lars B, \mi{not}\ \alpha'$ and $\alpha' \lars B, \mi{not}\, \alpha$, where $\alpha'$ is a
fresh atom.

A rule $r$ resp.\ program $\Pi$ is \emph{ground}, if it is variable-free and  
a \emph{fact}, if in addition $n\,{=}\,0$.
Rules $r$ with variables 
stand for the sets $grd(r)$ of their ground instances; 
semantically, $\Pi$ induces a set $\AS(\Pi)$ of stable models (answer sets)
\cite{gelf-lifs-88}
which
are Herbrand models (i.e., sets $I$ of ground atoms) of $\Pi$ 
justified
by the rules, 
in that $I$ is a $\subseteq$-minimal model of $f\Pi^I=$ $\{ r \in grd(\Pi)
\mid I \models B(r)\}$ \cite{FLP04}, where $grd(\Pi) = \bigcup_{r\in \Pi}grd(r)$. 
A program $\Pi$ is \emph{unsatisfiable}, if $\AS(\Pi)=\emptyset$.

The following rules show the part of
a Sudoku encoding that guesses an assignment of symbols to the cells, and ensures that each cell has a number.  
{\small
\begin{align}
\{sol(X,Y,N)\} \lars \mi{not}\ occupied(X,Y), num(N),\nonumber\\
\mi{row}(X),\mi{column}(Y).\nonumber\\
hasNum(X,Y) \lars sol(X,Y,N).\label{eq:sudokumain}\\
\lars \mi{not}\ hasNum(X,Y),\mi{row}(X),\mi{column}(Y).\nonumber
\end{align}
}
Further constraints are to ensure that cells in the same column \eqref{eq:sudokucol} or same row  \eqref{eq:sudokurow} do not contain the same symbol. 
{\small
\begin{align}
\lars sol(X,Y_1,M), sol(X,Y_2,M), Y_1 < Y_2.\label{eq:sudokucol}\\
\lars sol(X_1,Y,M), sol(X_2,Y,M), X_1 < X_2.\label{eq:sudokurow} 
\end{align}
}
An additional constraint (omitted due to space) ensures that the cells in the same sub-region also satisfies this condition.

\leanparagraph{Domain Abstraction} 
The generic notion of abstraction for ASP is as
follows:
\begin{defn}[\cite{zgskr18}]
\label{defn:abstraction}
Let $\Pi$, $\Pi'$
be two ground programs on sets $\Lits$, $\Lits'$ of atoms,
respectively,  where $|\Lits| \geq |\Lits'|$. 
Then $\Pi'$ is an \emph{abstraction} of $\Pi$,
if some mapping $m \,{:}\, \Lits \rightarrow \Lits'$
exists such that for each $I\,{\in}\, \AS(\Pi)$,
$I' = \{m(a) \,{\mid}\, a \,{\in}\, I\}$ is an answer set of $\Pi'$.
\end{defn}

For non-ground programs $\Pi$ with domain (Herbrand universe) $D$, \cite{zgs19jelia} introduced
\emph{domain abstraction mappings} $m\,{:}\,D\,{\rightarrow}\,
\widehat{D}$ for a set $\widehat{D}$ 
with $|\widehat{D}|\,{\leq}\,|D|$,  which 
divide $D$ into 
\emph{clusters} $\{ d \,{\in}\, D \,{\mid}\, m(d) \eqs \hat{d}\}$ of
elements
seen as equal.
Any 
such mapping $m$ naturally extends to 
the Herbrand base $\Lits\,{=}\,\mathit{HB}_\Pi$ of $\Pi$ by
$m(p(c_1,\ldots,c_n))\,{=}\, p(m(c_1),\ldots,m(c_n))$. 
E.g., for 
a graph coloring problem with an instance $\mi{node}(a)$, $\mi{node}(b)$, $\mi{node}(c)$ and $edge(a,b)$, an abstraction over the nodes such as $a \,{\mapsto}\, \hat{a}$, $\{b,c\} \,{\mapsto}\, \hat{b}$ (denoted $\{\{a\}/\hat{a},\{b,c\}/\hat{b}\}$) means to abstract over the node constants and obtain $\mi{node}(\hat{a})$, $\mi{node}(\hat{b})$, $\mi{edge}(\hat{a},\hat{b})$. An abstract program is then constructed that achieves an abstraction over the abstract atoms.
 
To build 
an abstract (non-ground) program $\Pi^m$, 
the procedure in  \cite{zgs19jelia} focuses on rules of form $r: l \leftarrow B(r), \Gamma_{\rel}(r)$ where the variables in $B(r)$ are standardized apart
and $\Gamma_{\rel}$ consists
of binary atoms $\rel(X,c)$ or $\rel(X,Y)$ 
on built-ins $\rel$ (e.g., $=$, $<, \leq, \neq$) that constrain the variables in $B(r)$. 
It abstracts each rule by 
treating the uncertainties caused by 
the domain abstraction $m$. 
To lift a built-in relation $\rel$, a set $\Reltype_m$ of atoms is
computed that distinguishes the cases for the truth value of $\rel$ in
the abstract domain,
which are respected during abstract rule construction. 
A non-ground program $\Pi^m$ is constructed such that 
for every $I\,{\in}\,\AS(\Pi)$, 
$\widehat{I}{\eqs}m(I)\,{\cup}\,\Reltype_m$ is an answer set of
$\Pi^m$,
where
$m(I) = \{ m(\alpha) \mid \alpha \in I \}$. 
In general, an \emph{over-approximation} of $\Pi$ is achieved, 
i.e., 
an answer set $\widehat{I}$ of $\Pi^m$ 
may not have a corresponding original answer set;
$\widehat{I}$ is \emph{concrete}, if $\widehat{I}\eqs m(I)\cups \Reltype_m\!$
for some $I{\ins}\AS(\Pi)$, else it is \emph{spurious}.
If an abstract answer set
$\widehat{I}$ is spurious, 
one can either
compute some other abstract answer set(s) and check
for concreteness, or {\em refine}\/ the mapping $m$, 
by dividing the abstract clusters to a finer grained domain.

\section{Abstracting Domain Relations}

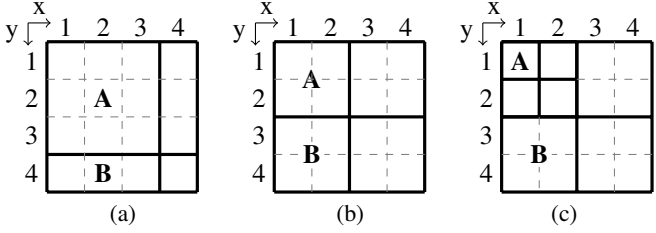
\begin{figure}[t!]
\caption{Abstractions over grid-cells}
\label{fig:gridabs}
\centering%
\begin{subfigure}[t]{0.17\textwidth}
 \begin{tikzpicture}[scale=0.5]
  \useasboundingbox (-1,0) rectangle (5,4.8);

  \begin{scope}
    \draw[step=4cm,very thick, scale=1] (0, 0) grid (4, 4);
\node at (1.5,2.5) {\textbf{A}};
\node[minimum size=0.5cm] at (0.5,4.4) {1};
\node[minimum size=0.5cm] at (1.5,4.4) {2};
\node[minimum size=0.5cm] at (2.5,4.4) {3};
\node[minimum size=0.5cm] at (3.5,4.4) {4};
\node[minimum size=0.5cm] at (-0.4,3.5) {1};
\node[minimum size=0.5cm] at (-0.4,2.5) {2};
\node[minimum size=0.5cm] at (-0.4,1.5) {3};
\node[minimum size=0.5cm] at (-0.4,0.5) {4};

\draw[very thick]  (3,4) -- (3,0); 
\draw [dashed,color=gray] (2,4) -- (2,0); 
\draw [very thick] (0,1) -- (4,1); 
\draw [dashed,color=gray] (1,4) -- (1,0);
\draw [dashed,color=gray] (0,2) -- (4,2);
\draw [dashed,color=gray] (0,3) -- (4,3);  
\node[circle,inner sep=0pt,fill=white] at (1.5,0.5) {\textbf{B}};

\draw [->] (-0.5,4.5) -- node[align=center, above] {x}(0.1,4.5);
\draw [->] (-0.5,4.5) -- node[align=center, left] {y}(-0.5,3.9);
\node at (5.3,2) {~};
  \end{scope}
 \end{tikzpicture}%
\caption{}
\label{fig:toyabs1}%
\end{subfigure}%
\begin{subfigure}[t]{0.17\textwidth}
\begin{tikzpicture}[scale=0.5]
  \useasboundingbox (-1,0) rectangle (5,4.8);

  \begin{scope}
    \draw[step=2cm,very thick, scale=1] (0, 0) grid (4, 4);
\node[circle,inner sep=0pt,fill=white] at (1,3) {\textbf{A}};
\node[minimum size=0.5cm] at (0.5,4.4) {1};
\node[minimum size=0.5cm] at (1.5,4.4) {2};
\node[minimum size=0.5cm] at (2.5,4.4) {3};
\node[minimum size=0.5cm] at (3.5,4.4) {4};
\node[minimum size=0.5cm] at (-0.4,3.5) {1};
\node[minimum size=0.5cm] at (-0.4,2.5) {2};
\node[minimum size=0.5cm] at (-0.4,1.5) {3};
\node[minimum size=0.5cm] at (-0.4,0.5) {4};
\draw [dashed,color=gray] (3,4) -- (3,0); 
\draw [dashed,color=gray] (0,1) -- (4,1); 
\draw [dashed,color=gray] (1,4) -- (1,0);
\draw [dashed,color=gray] (0,3) -- (4,3);  
\node[circle,,inner sep=0pt,fill=white] at (1,1) {\textbf{B}};

\draw [->] (-0.5,4.5) -- node[align=center, above] {x}(0.1,4.5);
\draw [->] (-0.5,4.5) -- node[align=center, left] {y}(-0.5,3.9);
\node at (5.3,2) {~};

  \end{scope}
 \end{tikzpicture}%
\caption{}
\label{fig:toyabs2}%
\end{subfigure}%
\begin{subfigure}[t]{0.15\textwidth}
\begin{tikzpicture}[scale=0.5]
  \useasboundingbox (-1,0) rectangle (5,4.8);

  \begin{scope}
    \draw[step=2cm,very thick, scale=1] (0, 0) grid (4, 4);
    \draw[step=1cm,very thick, scale=1,yshift=2cm,xshift=0cm] (0, 0) grid (2, 2);
\node at (0.5,3.5) {\textbf{A}};
\node[minimum size=0.5cm] at (0.5,4.4) {1};
\node[minimum size=0.5cm] at (1.5,4.4) {2};
\node[minimum size=0.5cm] at (2.5,4.4) {3};
\node[minimum size=0.5cm] at (3.5,4.4) {4};
\node[minimum size=0.5cm] at (-0.4,3.5) {1};
\node[minimum size=0.5cm] at (-0.4,2.5) {2};
\node[minimum size=0.5cm] at (-0.4,1.5) {3};
\node[minimum size=0.5cm] at (-0.4,0.5) {4};
\draw [dashed,color=gray] (3,4) -- (3,0); 
\draw [dashed,color=gray] (0,1) -- (4,1); 
\draw [dashed,color=gray] (1,2) -- (1,0);
\draw [dashed,color=gray] (2,3) -- (4,3);  
\node[circle,,inner sep=0pt,fill=white] at (1,1) {\textbf{B}};

\draw [->] (-0.5,4.5) -- node[align=center, above] {x}(0.1,4.5);
\draw [->] (-0.5,4.5) -- node[align=center, left] {y}(-0.5,3.9);
\node at (5.3,2) {~};

  \end{scope}
 \end{tikzpicture}
\caption{}
\label{fig:toyabs}%
\end{subfigure}
\nbls
\end{figure}

In order to obtain an abstraction over a grid-cell that allows to adjust its granularity, multi-dimensionality has to be considered, which can not be handled by \cite{zgs19jelia}.

\leanparagraph{Need for multi-dimensionality} Consider the abstractions in Fig.~\ref{fig:gridabs}. Achieving those in Figs.~\ref{fig:toyabs1}-\ref{fig:toyabs2} is possible by a mapping over the rows and columns independently such as $m_{row}\,{=}\,$ $m_{col}\,{=}\,\{\{1,2,3\}/a_{1,2,3},\{4\}/a_4\}$ and $m_{row}\,{=}$ $\,m_{col}\,{=}\,\{\{1,2\}/a_{1,2},\{3,4\}/a_{3,4}\}$. For a given program $\Pi$, one can construct the abstract program $(\Pi^{m_{row}})^{m_{col}}$ using the method in \cite{zgs19jelia}. However to achieve Fig.~\ref{fig:toyabs}, rows and columns must be \emph{jointly}\/ abstracted. While the cells $(a_i,b_j), 1\,{\leq}\,i,j\,{\leq}\, 2$ are
singletons mapped from $(i,j)$, the
other abstract regions
are only given
by
\beq
m_{row,col}(x,y) = \left\{ 
\ba{@{}l@{\hspace{-1.1em}}l@{}}
(a_{12},b_{34}) & \qquad x \in\{1,2\} ,y \in \{3,4\} \\
(a_{34},b_{12}) & \qquad x \in\{3,4\} ,y \in \{1,2\} \\
(a_{34},b_{34}) & \qquad x \in\{3,4\} ,y \in \{3,4\} \\
\ea\right.
\eeq {eq:m2}
Observe that the abstract row $a_{12}$ describes a cluster that abstracts over the individual abstract rows $a_1,a_2$. The original rows $\{1,2\}$ are mapped to $\{a_{12}\}$ only in combination with columns $\{3,4\}$, otherwise they are mapped to $\{a_1,a_2\}$.

\leanparagraph{Undefined lifted relations} Consider the rule \eqref{eq:sudokucol} standardized apart over rows and columns, to have the relations $X_1=X_2$ and $Y_1 < Y_2$. For the mapping $m_{row,col}$ (Fig.~\ref{fig:toyabs}), if these relations are lifted by following \cite{zgs19jelia}, although the relation over the y-axis is still
defined (as $A$ is located above of $B$), i.e., $A_Y \leq B_Y$, $A_X = B_X$ is unclear as the abstract clusters for $X$ values are different due to different levels of abstraction.




Before introducing 
domain mappings over multiple subdomains (\emph{sorts}), i.e.,
$m \,{:}\, D_1 \,{\times}\, \dots \,{\times}\, D_n \rightarrow \hat{D}_1 \,{\times}\, \dots \,{\times}\, \hat{D}_n$,
we first 
deal with relations over different levels of abstraction.
For this, we briefly introduce an alternative approach for constructing abstract programs
by abstracting the relations in
the style of existential abstraction \cite{clarke03}. 
\nop{Semantically,
the constructed abstract program is the same as the one in
\cite{zgs19jelia}, but it
allows 
for more powerful abstraction mappings.
}

\subsection{Abstract Relations}

An abstract relation $\absrel$ for a relation $\mi{rel}$ is as follows:
\nqbls
{\small
\begin{align}
(\forall\hat{d}_i\in\widehat{D})
{\absrel}(\hat{d}_1,\dots,\hat{d}_k) \Leftrightarrow& \exists x_i
\in m^{-1}(\hat{d}_i).\mi{rel}(x_1,\dots,x_k).\nonumber\\
(\forall\hat{d}_i\in\widehat{D})
{\mi{neg}\_\absrel}(\hat{d}_1,\dots,\hat{d}_k) \Leftrightarrow& \exists x_i \in m^{-1}(\hat{d}_i).\neg\mi{rel}(x_1,\dots,x_k).\nonumber
\end{align}
}
\nbls

\noindent
Abstract relation ${\absrel}(\hat{d}_1,\dots,\hat{d}_k)$ (resp. ${\mi{neg}\_\absrel}(\hat{d}_1,\dots,\hat{d}_k)$)
is true
if for some original values $\mi{rel}$ holds (resp. does not hold);
$\neg{\absrel}(\hat{d}_1,\dots,\hat{d}_k)$ (resp. $\neg{\mi{neg}\_\absrel}(\hat{d}_1,\dots,\hat{d}_k)$)
is true otherwise.
Notably, both versions ${\absrel}$ and ${\mi{neg}\_\absrel}$ may hold
simultaneously, depending on the abstract domain clusters.
For example, for the mapping $m_{row,column}$ (Fig.~\ref{fig:toyabs}), the abstract relation $Y \widehat{<} Y_1$ holds true, since $Y_1 {<} Y_2$ for all $Y_1$ and $Y_2$ mapped to $A$ and $B$, resp. The abstract relation $X_1 \widehat{=} X_2$ and its negation both hold true, since $X_1{=}X_2$ holds only for some $X_1$ and $X_2$ values mapped to $A$ and $B$, resp.

Notice that having both $rel$ and $neg\_rel$ hold means an uncertainty on the truth value of the relation in the abstract clusters. This brings us to determining the types of the relations over the abstract clusters.

\leanparagraph{Abstract relation types}
For the abstract relation predicates $\widehat{re}l(\hat{d}_1,\dots,\hat{d}_k)$ and $\mi{neg}\_\widehat{\mi{re}}\mi{l}(\hat{d}_1,\dots,\hat{d}_k)$, the following cases $\reltype{I}-\reltype{III}$ occur in a mapping:

\vspace*{-\baselineskip}

{\small
\beeq
\begin{array}{@{~~}l@{:\ }r@{\ \wedge\ } r@{}}
\reltype{I}^{\absrel}(\hat{d}_1,\dots,\hat{d}_k)
  & \absrel(\hat{d}_1,\dots,\hat{d}_k)
    & \neg\mi{neg}\_\absrel(\hat{d}_1,\dots,\hat{d}_k) \\
\reltype{II}^{\absrel}(\hat{d}_1,\dots,\hat{d}_k)
  & neg\_\absrel(\hat{d}_1,\dots,\hat{d}_k)
    & \neg\absrel(\hat{d}_1,\dots,\hat{d}_k)\\
\reltype{III}^{\absrel}(\hat{d}_1,\dots,\hat{d}_k)
  & \absrel(\hat{d}_1,\dots,\hat{d}_k) 
    & \mi{neg}\_\absrel(\hat{d}_1,\dots,\hat{d}_k)
\end{array}
\eeeq
}
\nhbls

Type I is the case where the abstraction does not cause uncertainty
for the relation, thus the rules that contain $\widehat{re}l$ with
type I can remain the same in the abstract program. Type II shows the
cases where $\widehat{re}l$ does not hold in the abstract
domain. Type III is the uncertainty case, which needs to be dealt with
when creating the abstract rules. 
Note that definitions of Types I and II are similar to the case of lifted relations \cite{zgs19jelia}, while Type III corresponds to all uncertainty cases for lifted relations.
For an abstraction $m$,
we compute the set $\Reltype_m$ of all atoms 
$\reltype{\iota}^{\absrel}(\hat{d}_1,\dots,\hat{d}_k)$
where $\iota\,{\in}\, \{\mathrm{
I,II,III}\}$ is the type of $\absrel(\hat{d}_1,\dots,\hat{d}_k)$
for $m$.

\subsection{Abstraction Procedure}

For ease of presentation, we consider programs with rules having (i) a single relation atom, and (ii) no cyclic dependencies between non-ground atoms. Removing these restrictions can be easily done as in \cite{zgs19jelia}.

%
\begin{defn}[rule abstraction]
\label{defn:absrule}
Given a rule $r:\ l \leftarrow B(r),$ $\rel(t_1,\dots,t_k)$ and a domain mapping $m$, the set $r^m$ contains the following rules.

\be[\quad(a)]
\item $m(l) \leftarrow m(B(r)), \reltype{I}^{\absrel}(\hat{t}_1,\dots,\hat{t}_k).$
\item 
$\{m(l)\} \leftarrow m(B(r)), \reltype{III}^{\absrel}(\hat{t}_1,\dots,\hat{t}_k).$ 
\nop{
\item For $l_i \in S^-_{\rel}(r)$:\\[-1pt]
$\{m(l)\} \leftarrow m(B_{l_i}^{sh}(r)), \reltype{III}^{\widehat{\rel}}(\hat{t}_1,\dots,\hat{t}_k), \mi{isCluster}(l_i).$
}
\item For all $L \subseteq B^{-}(r)$, $l_i \in L$ and $j \in arg(l_i)$:\\
$\{m(l)\} \lars m(B^{\myshift}_{L}(r)),\reltype{I}^{\widehat{\rel}}(\hat{t}_1,\dots,\hat{t}_k),\mi{isCluster}(\hat{j}).$\\
$\{m(l)\} \lars m(B^{\myshift}_{L}(r)),\reltype{III}^{\widehat{\rel}}(\hat{t}_1,\dots,\hat{t}_k),\mi{isCluster}(\hat{j}).$
\ee 

\end{defn}
\noindent  
where $\!B^{\myshift}_{L}(r){=}B^+(r)\,{\cup}\, L,\mi{not}\, B^-(r){\setminus} L$;
the auxiliary atom $\mi{isCluster}(\hat{d})$ 
is  true for proper   (non-singleton) clusters
$\hat{d}$.

The idea is to introduce guesses when there is an uncertainty over the relation holding in the abstract domain $(b)$, or over the negated atoms due to the abstract clusters $(c)$ (by considering all combinations of the negative literals), and otherwise just abstracting the rule $(a)$.

We construct $\Pi^m$ modularly, rule by rule, and obtain:

\begin{thm}
Let $m$ be a domain mapping of a program $\Pi$.
For every $I\,{\in}\,\AS(\Pi)$, 
$\widehat{I}{\eqs}m(I)\,{\cup}\,\Reltype_m$ is an answer set of $\Pi^m$.
\end{thm}

The introduced approach may construct a program with more spurious abstract answer sets than the one obtained with \cite{zgs19jelia}. An additional rule can be added for step $(b)$ to avoid too many spurious guesses, and reach the same answer sets as in \cite{zgs19jelia}. Choice rules, i.e., $\{l\}\lars B$, are treated by keeping the choice in the head.

The use of abstract relations puts no restriction on their form, and thus opens a wide-range of possible applications. 

\section{Multi-Dimensional Abstraction}

Abstracting over a set of sorts in the domain by adhering to a structure can be done using an abstraction mapping in the form $m \,{:}\, D_1 \,{\times}\, \dots \,{\times}\, D_n \rightarrow \hat{D}_1 \,{\times}\, \dots \,{\times}\, \hat{D}_n$. If a rule has relations
over these sorts, a joint abstract relation and its types must be computed (see Appendix~\ref{app:joint}). In order to ensure that the rules in $\Pi^m$ consider valid abstracted sorts, the occurrences of these sort names need to be replaced with a new \emph{object} name.
For example, the abstract program for Sudoku \eqref{eq:sudokumain}-\eqref{eq:sudokurow}, where the occurrences of $\mi{row}(X)$, $\mi{column}(Y)$ are replaced by $cell(X,Y)$, is as follows.

{\small
\begin{align*}~ \\[-2\baselineskip]
hasNum(X,Y) \lars sol(X,Y,N).\\
\{sol(X,Y,N)\} \lars \mi{not}\ occupied(X,Y), num(N), cell(X,Y).\\
\{sol(X,Y,N)\} \lars occupied(X,Y), num(N), isCluster(X).\\
\{sol(X,Y,N)\} \lars occupied(X,Y), num(N), isCluster(Y).\\
\lars \mi{not}\ hasNum(X,Y), cell(X,Y).\\
\lars sol(X_1,Y_1,M), sol(X_2,Y_2,M), relr3(X_1,Y_1,X_2,Y_2,i).\\
\lars sol(X_1,Y_1,M), sol(X_2,Y_2,M), relr4(X_1,Y_1,X_2,Y_2,i).
\end{align*}
\nbls
}

\leanparagraph{Quad-tree Abstraction}%
We consider a generic quad-tree representation 
for a systematic refinement of abstractions on grid-cell environments (a concept used in path planning \cite{kambhampati1986multiresolution}). Initially, an environment may be abstracted to 4 regions of $n/2 \times n/2$ grid-cells each. This amounts to a
tree with 4 leaf nodes that correspond to the main regions, with level $log_2(n)$. Each region then contains 4 leaves of smaller regions. A refinement of a region then amounts to dividing the region into 4, i.e., expanding the representing leaf with its four leaves. The leaves of the main quad-tree are then the original cells of the grid-cell. 

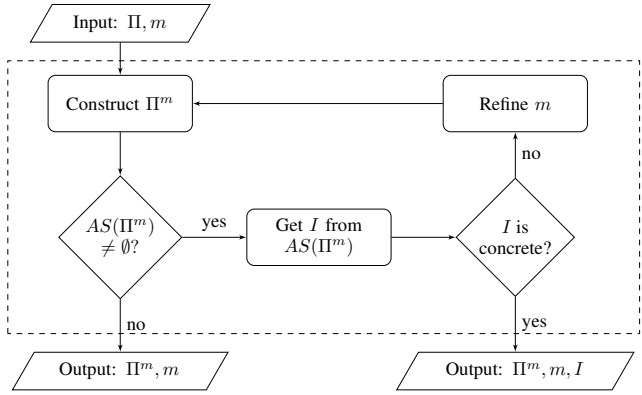
\begin{figure}[t!]
\caption{Abstraction \& Refinement Methodology} 
\label{fig:absrefmet}
\nhbls
\centering
\tikzstyle{decision} = [diamond, draw, 
    text width=4.5em, text badly centered, node distance=2.5cm, inner sep=0pt]
\tikzstyle{block} = [rectangle, draw,, 
    text width=7em, text centered, rounded corners, minimum height=3em]
\tikzstyle{line} = [draw, -latex']
\tikzstyle{cloud} = [draw, ellips, node distance=3cm,
    minimum height=2em]
 \tikzstyle{data} =[draw,trapezium,trapezium left angle=60,trapezium right angle=-60,minimum height=2em,text width=7em]
    
\resizebox{0.48\textwidth}{!}{
\begin{tikzpicture}[node distance = 1.5cm, auto]
    \node [data,text width=5em] (init) {Input: $\Pi, m$};
    \node [block, below of=init] (construct) {Construct $\Pi^m$};    
    \path [line] (init) -- (construct);
\draw[draw=black,dashed] (-2.1,-5.8) rectangle ++(11.8,5.1);
\node [decision, below of=construct] (decide) {$AS(\Pi^m)$\\$\neq \emptyset?$};
    \node [data,text width=6.5em, below of=decide, node distance=2.5cm] (stop1) {Output: $\Pi^m, m$};
    
    \node [block, right= 1.2cm of decide, node distance=2.5cm] (getanswerset) {Get $I$ from $AS(\Pi^m)$};
    \node [decision, right= 1.2cm of getanswerset, node distance=1.5cm] (isconcrete) {$I$ is concrete?};
   \node [data, below of=isconcrete, node distance=2.5cm,text width=7.4em] (stop2) {Output: $\Pi^m, m, I$};
    \node [block, above of=isconcrete, node distance=2.5cm] (refine) {Refine $m$};
    \path [line] (construct) -- (decide);
    \path [line] (refine) -- (construct);
    \path [line] (decide) -- node {yes} (getanswerset);
    \path [line] (getanswerset) -- (isconcrete);
    \path [line] (decide) -- node {no}(stop1);
    \path [line] (isconcrete) -- node {yes}(stop2);
    \path [line] (isconcrete) -- node[right] {no}(refine);
\end{tikzpicture}
}
\nbls
\end{figure}

\section{Finding Abstract Solutions}

Figure~\ref{fig:absrefmet} shows the overall methodology. 
Deciding on a refinement is done using the hints obtained from correctness checking
of an abstract answer set.
Correctness checking depends on the following characteristic of spuriousness.
\begin{prop}[cf.~\cite{zgs19jelia}]
\label{prop:query}
 $\widehat{I}$ is spurious iff $\Pi \cup Q_{\widehat{I}}^m$ is unsatisfiable, where
 $Q_{\widehat{I}}^m$ represents the constraints 

 \smallskip
 
\noindent
\hfill$\begin{array}{@{}l@{\quad}l@{\quad}r@{}}
 \lars \{\alpha \mids m(\alpha)\eqs\hat{\alpha}\}\leq 0.
  &\hat{\alpha} \ins \widehat{I} \setminuss \Reltype_m
		&\equationnumberhere\label{eqn:check1}\\ 
 \lars \alpha.
  &\hat{\alpha} \notins \widehat{I} \setminuss \Reltype_m, m(\alpha)\eqs\hat{\alpha}
		&\equationnumberhere\label{eqn:check2}
\end{array}$
\smallskip
\end{prop}
Here  \eqref{eqn:check1} ensures that a witnessing answer set $I$ of
$\Pi$ 
contains
for every non-$\reltype{\iota}$, abstract atom 
in $\widehat{I}$
some atom that is mapped to
it. The constraint \eqref{eqn:check2} ensures that $I$
has no atom that is mapped to an abstract atom not in
$\widehat{I}$.

For a spurious abstract answer set $\widehat{I}$, using Proposition~\ref{prop:query} to check its correctness returns unsatisfiability without a reason. To obtain hints for refinement of the abstraction, in \cite{zgs19jelia} we proposed a debugging approach to be used during this check to obtain the reason for unsatisfiability. Since the proposed approach was unable to address all debugging cases, we extend it to a more sophisticated method  combining the debugging notions in \cite{brain2007debugging} with the meta-program constructed by \cite{oetsch2010catching}.

\subsection{Implementation}
\label{sec:imp}
The implementation mDASPAR
\footnote{\label{fn:1}\url{http://www.kr.tuwien.ac.at/research/systems/abstraction/}. 
The problem encodings and all user explanations can be found in \url{./mdaspar_material.zip}}
uses Python, Clingo and the meta-program constructer of \cite{oetsch2010catching}. 
\begin{algorithm}[t!]
\caption{Abstraction\&Refinement}
\label{alg:absref}
\footnotesize
 \KwIn{$\Pi$, $m_{init}$, $S$ (set of abstracted sorts),} 
 \KwOut{$\Pi^m, m_{final},I$}
 $m=m_{init}$;\\
 $\Pi^m=constructAbsProg(\Pi,S)$;\label{line:abs}\\
 ${\cal T}_m=computeRelTypes(\Pi,m)$;\label{line:rel}\\
 $\Pi_{debug}=constructDebugProg(\Pi,S)$;\label{line:debug}\\
 \While{$AS(\Pi^m,{\cal T}_m)\neq\emptyset$}{
 $A = \mi{getAnsSets}(\Pi^m,{\cal T}_m)$;\label{line:diverse}\\
 $C_{list}=[]$;\\
 \For{$I \in A$}{
 	$C=checkCorrectness(\Pi_{debug},m,I)$;\label{line:ab}\\
 	\If(\tcc*[h]{ $I$ concrete}){$C|_{ab} = \emptyset$}{
 		\Return{$\Pi^m,m,I$}\label{line:concrete}
 	}
 	\Else{
 		$C_{list}.append(C)$;
 	}
 }
 $m=decideRefinement(m,C_{list})$;\label{line:refine}\\
 ${\cal T}_m=computeRelTypes(\Pi,m)$;\\
 }
 \Return{$\Pi^m,m,\emptyset$}\label{line:end}
\end{algorithm}
The procedure for the abstraction and refinement method (following Fig.~\ref{fig:absrefmet}) is shown in
Algorithm~\ref{alg:absref}. For the constructed program $\Pi^m$, a set $A$ of abstract answer sets is computed (Line~\ref{line:diverse}) and each such $I$ is
checked for concreteness (Line~\ref{line:ab}) using the constructed debugging program
$\Pi_{debug}$. The collection $C_{list}$ of 
the debugging outputs from checking the answer sets in $A$ 
is used to decide on a
refinement over the mapping $m$ (Line~\ref{line:refine}). The debugging outputs consist of inferred $\mi{refine}$ atoms that contain domain elements with non-singleton clusters. 
The cluster with the largest number of $\#\mi{refine}$ atoms/size of cluster is picked to refine.

\leanparagraph{Incremental checking}
To see the effect of 
different ways of
correctness checking, 
we considered in addition to 
default 
debugging three approaches:
\begin{list}{~}{
\setlength{\topsep}{2pt}
\setlength{\itemsep}{0pt}
\setlength{\leftmargin}{0pt}
\setlength{\itemindent}{0pt}}
\item[] {\bf(2-phase)} First using a simplified debugging program to
  distinguish the rules/atoms that cause issues, and then using the
  output to steer the full debugging program towards the abstract
  elements involved.
\item[] {\bf(time-inc)} For problems 
with a clear order on the atoms
for the solution (e.g., in ASP planning encodings action atoms
have \emph{time} arguments), we check incrementally over
the order. 
\item[] {\bf (grid-inc)} If no such clear order exists, we check 
via
incrementally con\-cretizing the abstract domain. If the abstract answer
set is concrete 
wrt.\ a partially concretized abstraction,
the concretization is increased to 
redo the check.
\end{list}

\section{Evaluation: Unsolvable Problem Instances}

We investigated getting explanations of unsatisfiable
grid-cell problems 
by achieving an abstraction over the instance to focus on the
troubling area. In addition to Reachability (R) and Sudoku (S), we
have the following benchmark problems:%
\footnotemark[1]
\smallskip

\noindent \emph{Knight's Tour} (KT), which is finding a tour on which a knight
visits each square of a board once and returns to the starting
point.
It is commonly used in ASP Competitions,
with possible addition of forbidden cells.
This problem is encoded by guessing a set of $\mi{move}(X_1,Y_1,X_2,Y_2)$ atoms and ensuring that each cell has only one incoming and one outgoing movement.
There is no \emph{time} sort as in planning.
\smallskip

\noindent \emph{Visitall}, which is the planning problem of visiting every cell (without revisiting a cell) extended with obstacles. Visitall 
is encoded in two forms: (V) as a planning problem, to find a sequence of actions that visits every cell, or (V$_{\textup{KT}}$) as a combinatorial problem similar to the Knight's Tour encoding.

\begin{table}[t!]
\caption{Evaluation results}
\label{fig:results_unsatgrid}
\nhbls
\centering
\small
\renewcommand{\arraystretch}{1.1}
\begin{tabular}{|l@{~}|@{~}c@{~}|c@{~~~}c|c@{~~~}c||c@{~~~}c|}\hline
& ~debugging~ & \multicolumn{2}{c|}{average} & \multicolumn{2}{c||}{minimum} & \multicolumn{2}{c|}{best}  \\
& ~type~ & steps & cost & steps &  cost & steps & cost  \\
\hline
\hline
\multirow{2}{*}{R}& default & 5.4&0.227 & 5.4&0.227 & \multirow{2}{*}{5.0}&\multirow{2}{*}{0.208} \\
\cline{2-6}
& 2-phase & 5.5&0.233 & 5.3&0.222 & &\\
 \hline
\hline
\multirow{2}{*}{S}& default & 6.5& 0.696& 5.1& 0.550& \multirow{2}{*}{3.2}&\multirow{2}{*}{0.371} \\
\cline{2-6}
& 2-phase & 4.3&0.476 & 3.4&0.391 & & \\
\hline
\hline
\multirow{2}{*}{KT}& 2-phase &14.3&0.643& 10.4&0.460 & \multirow{2}{*}{5.6} &\multirow{2}{*}{0.245}\\ 
\cline{2-6}
& grid-inc & 10.1&0.442&6.3&0.277 &&\\
\hline
\hline
\multirow{2}{*}{V}&2-phase & 16.2& 0.708& 13.9&0.608  & \multirow{2}{*}{8.7} & \multirow{2}{*}{0.360}\\
\cline{2-6}
& time-inc &16.3& 0.712& 13.5&0.569  &  & \\
\hline
\hline
\multirow{2}{*}{V$_{\textup{KT}}$} & 2-phase& 15.7 & 0.693& 13.0&0.572 & \multirow{2}{*}{7.6} & \multirow{2}{*}{0.317}\\
\cline{2-6}
 & grid-inc & 13.0& 0.569& 10.3&0.449 &&\\
\hline
\end{tabular}
\nhbls
\end{table}

\leanparagraph{Measuring abstraction quality}
We consider a \emph{quality measure} of the quad-tree abstraction by normalizing the number of abstract regions of a certain size and their level in the quadtree. The cost of a mapping $m$ over an $n \,{\times}\, n$ grid is
\[
\textstyle c(m) ={\sum_{i=0}^{\ell}}\,r_{2^i}(m) (\ell\,{-}\,i) \;\big/\; {\sum_{i=0}^{\ell}}\; n^22^{-i^2}(\ell\,{-}\,i),
\]
where $\ell=\log_2(n)\,{-}\,1$,
$r_{2^i}(m)$ is the number of abstract regions of size $2^i
\,{\times}\, 2^i$ in $m$, and ${n^2}{2^{-i^2}}$ 
is the number of 
abstract regions of size $2^i \,{\times}\, 2^i$ in the $n \,{\times}\,
n$-sized cell. The factor $\ell{-}i$ is a weight that gives higher cost to abstractions with more low-level regions. The abstraction mapping with the smaller cost is considered to be of \emph{better quality}.

\leanparagraph{Evaluation results}%
We generated 10 unsatisfiable instances for each benchmark, to compare different debugging approaches in terms of the average refinement steps and average costs of the resulting abstractions over 10 runs, and also on the best outcome obtained (with minimum refinement steps and minimum mapping cost) among the 10 runs. Table~\ref{fig:results_unsatgrid} shows the evaluation results.
The right-most two columns are
for checking the existence of a coarser abstraction from the best outcome obtained in the runs.
The time to find an optimal solution when debugging the
concreteness checking
was limited by 50 seconds. If none is found within the time limit, the refinement is decided on the basis of suboptimal analyses.

For Reachability and Sudoku, we observe that abstractions close to the best possible ones can be obtained. 
Abstractions that are slightly better were obtained with 2-phase debugging, due to putting the focus on the right part of the abstraction after the first step. For Knight's Tour and Visitall, we observe that incremental checking can obtain better abstractions. For 2-phase debugging, the program mostly had to decide on suboptimal concreteness checking outputs, due to timeouts. Additionally, for V, 2-phase debugging caused memory errors (when over 500 MB) on some runs for some instances, thus not all 10 runs could be conducted. 

We can also see a difference of the resulting abstractions for the different encodings of Visitall. The planning encoding achieves unsatisfiability with finer abstractions, in order to avoid the spurious guesses of action sequences.

\section{User Study on Unsatisfiability Explanations}

We were interested in checking whether the obtained abstractions match the intuition behind a human explanation.
For Reachability and
Visitall, finding the reason for 
unsolvability of an instance is possible by looking at the obstacle layout.
Thus, we conducted a user study for these problems in order
to obtain the regions that humans focus on to realize the unsolvability of the problem instance.

As participants, we had 10 PhD students of Computer Science in TU Wien. We asked them
to mark the area which shows the reason (if more than one exists mark with different colors) for having unreachable cells in the Reachability instances
and the reason for not finding a solution that visits all the cells in the Visitall instances. Explanations for 10 instances of each problem were collected\footnotemark[1]. We discuss the results for both problems by showing two of the responses (expected and unexpected) and the best abstraction obtained from mDASPAR when starting with the initial mapping.

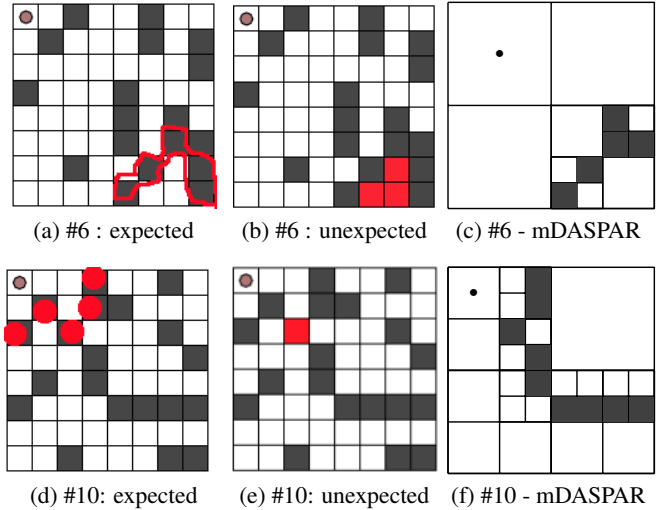
\begin{figure}[t!]
\caption{Explanations for unsolvable Reachability instances}
\label{fig:reachinsts}
\centering
\begin{subfigure}[t]{0.16\textwidth}
\centering
\includegraphics[scale=0.47,trim=0 0 0 0]{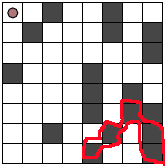}
\caption{\#6 : expected}
\label{fig:reachinst6-1}
\end{subfigure}
\begin{subfigure}[t]{0.16\textwidth}
\centering
\includegraphics[scale=0.47]{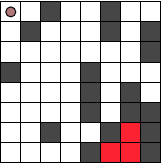}
\caption{\#6 : unexpected}
\label{fig:reachinst6-2}
\end{subfigure}
\begin{subfigure}[t]{0.15\textwidth}
\centering
\resizebox{!}{0.12\textheight}{%
\begin{tikzpicture}[scale=0.5]

  \begin{scope}
    \useasboundingbox (0,0) rectangle (8,8);
    \draw[step=4cm,thick, scale=1] (0, 0) grid (8, 8);
    \draw[step=1cm,thick, scale=1,yshift=2cm,xshift=6cm] (0, 0) grid (2, 2);
    \draw[step=2cm,thick, scale=1,xshift=4cm] (0, 0) grid (4, 4);
    \draw[step=1cm,thick, scale=1,yshift=0cm,xshift=4cm] (0, 0) grid (2, 2);
\node at (2,6) {$\bullet$};
\node[fill=darkgray,minimum size=0.47cm] at (4.5,0.5) {};
\node[fill=darkgray,minimum size=0.47cm] at (5.5,1.5) {};
\node[fill=darkgray,minimum size=0.47cm] at (6.5,2.5) {};
\node[fill=darkgray,minimum size=0.47cm] at (7.5,2.5) {};
\node[fill=darkgray,minimum size=0.47cm] at (6.5,3.5) {};
  \end{scope}
    
\end{tikzpicture}}
\caption{\#6 - mDASPAR}
\label{fig:reachinst6-3}
\end{subfigure}

\begin{subfigure}[t]{0.16\textwidth}
\centering
\includegraphics[scale=0.47,trim=1em 0 0 0]{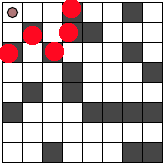}
\caption{\#10: expected}
\label{fig:reachinst10-1}
\end{subfigure}
\begin{subfigure}[t]{0.16\textwidth}
\centering
\includegraphics[scale=0.5]{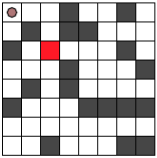}
\caption{\#10: unexpected}
\label{fig:reachinst10-2}
\end{subfigure}
\begin{subfigure}[t]{0.15\textwidth}
\centering
\resizebox{!}{0.12\textheight}{%
\begin{tikzpicture}[scale=0.5]

  \begin{scope}
    \useasboundingbox (0,0) rectangle (8,8);
    \draw[step=4cm,thick, scale=1] (0, 0) grid (8, 8);
    \draw[step=1cm,thick, scale=1,yshift=6cm,xshift=2cm] (0, 0) grid (2, 2);
    \draw[step=1cm,thick, scale=1,yshift=2cm,xshift=2cm] (0, 0) grid (2, 2);
    \draw[step=2cm,thick, scale=1] (0, 0) grid (4, 4);
    \draw[step=1cm,thick, scale=1,yshift=4cm,xshift=2cm] (0, 0) grid (2, 2);
    \draw[step=1cm,thick, scale=1,xshift=4cm,yshift=2cm] (0, 0) grid (2, 2);
    \draw[step=1cm,thick, scale=1,xshift=6cm,yshift=2cm] (0, 0) grid (2, 2);
    \draw[step=2cm,thick, scale=1,yshift=4cm] (0, 0) grid (4, 4);
    \draw[step=2cm,thick, scale=1,xshift=4cm] (0, 0) grid (4, 4);
\node at (1,7) {$\bullet$};
\node[fill=darkgray,minimum size=0.47cm] at (2.5,5.5) {};
\node[fill=darkgray,minimum size=0.47cm] at (3.5,6.5) {};
\node[fill=darkgray,minimum size=0.47cm] at (3.5,7.5) {};
\node[fill=darkgray,minimum size=0.47cm] at (3.5,4.5) {};
\node[fill=darkgray,minimum size=0.47cm] at (7.5,2.5) {};
\node[fill=darkgray,minimum size=0.47cm] at (6.5,2.5) {};
\node[fill=darkgray,minimum size=0.47cm] at (5.5,2.5) {};
\node[fill=darkgray,minimum size=0.47cm] at (4.5,2.5) {};
\node[fill=darkgray,minimum size=0.47cm] at (3.5,3.5) {};
  \end{scope}
    
\end{tikzpicture}}
\caption{\#10 - mDASPAR}
\label{fig:reachinst10-3}
\end{subfigure}
\end{figure}

\leanparagraph{Reachability}%
%
The expected explanations (e.g., Fig.s~\ref{fig:reachinst6-1} and \ref{fig:reachinst10-1}) focus on the obstacles that surround the unreachable cells, as they prevent them from being reachable. 
The explanation in Fig.~\ref{fig:reachinst6-2} puts the focus on the unreachable cells themselves,
and Fig.~\ref{fig:reachinst10-2} distinguishes a particular obstacle as a reason.
When the respective abstraction mappings are given to mDASPAR, it needs to refine further to distinguish more obstacles and achieve unsatisfiability. 
The mark in Figure~\ref{fig:reachinst10-2} is a possible solution to the unreachability of the cells, since removing the marked obstacle makes all the cells reachable. 

In ASP, checking whether all cells are reachable is straightforward, without introducing guesses. This is also observed to be helpful for mDASPAR, as most of the resulting abstractions were similar to the gathered answers. Since in the initial abstraction, the abstract program only knows that the agent is located in the upper-left abstract region, in instance \#10, mDASPAR follows a different path in refining the abstraction, and reaches the abstraction shown in Figure~\ref{fig:reachinst10-3}. Although not the same as the one given by the users, this abstraction also shows a reason for having unreachable cells.

\begin{figure}[t!]
\caption{Explanations for unsolvable Visitall instances}
\label{fig:visitinsts}
\centering
\begin{subfigure}[t]{0.16\textwidth}
\centering
\includegraphics[scale=0.5,trim=1em 0 0 0]{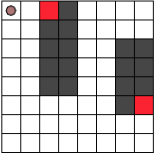}
\caption{\#1: expected}
\label{fig:visitinst1-1}
\end{subfigure}
\begin{subfigure}[t]{0.16\textwidth}
\centering
\includegraphics[scale=0.63]{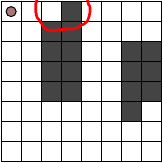}
\caption{\#1: unexpected}
\label{fig:visitinst1-2}
\end{subfigure}
\begin{subfigure}[t]{0.15\textwidth}
\centering
\resizebox{!}{0.12\textheight}{%
\begin{tikzpicture}[scale=0.5]

  \begin{scope}
    \useasboundingbox (0,0) rectangle (8,8);
    \draw[step=4cm,thick, scale=1] (0, 0) grid (8, 8);
    \draw[step=1cm,thick, scale=1,yshift=6cm] (0, 0) grid (2, 2);
    \draw[step=1cm,thick, scale=1,yshift=6cm,xshift=2cm] (0, 0) grid (2, 2);
    \draw[step=1cm,thick, scale=1,yshift=2cm,xshift=6cm] (0, 0) grid (2, 2);
    \draw[step=2cm,thick, scale=1,yshift=4cm] (0, 0) grid (4, 4);
    \draw[step=2cm,thick, scale=1,xshift=4cm] (0, 0) grid (4, 4);
    \draw[step=1cm,thick, scale=1,yshift=0cm,xshift=6cm] (0, 0) grid (2, 2);
\node at (0.5,7.5) {$\bullet$};
\node[fill=darkgray,minimum size=0.47cm] at (2.5,6.5) {};
\node[fill=darkgray,minimum size=0.47cm] at (3.5,6.5) {};
\node[fill=darkgray,minimum size=0.47cm] at (3.5,7.5) {};
\node[fill=darkgray,minimum size=0.47cm] at (6.5,2.5) {};
\node[fill=darkgray,minimum size=0.47cm] at (6.5,3.5) {};
\node[fill=darkgray,minimum size=0.47cm] at (7.5,3.5) {};
  \end{scope}
    
\end{tikzpicture}}
\caption{\#1 - mDASPAR}
\label{fig:visitinst1-3}
\end{subfigure}

\begin{subfigure}[t]{0.16\textwidth}
\centering
\includegraphics[scale=0.5,trim=1em 0 0 0]{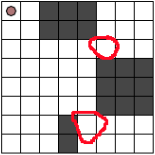}
\caption{\#10: expected}
\label{fig:visitinst10-1}
\end{subfigure}
\begin{subfigure}[t]{0.16\textwidth}
\centering
\includegraphics[scale=0.63]{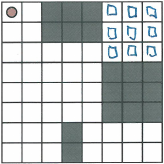}
\caption{\#10: unexpected}
\label{fig:visitinst10-2}
\end{subfigure}
\begin{subfigure}[t]{0.15\textwidth}
\centering
\resizebox{!}{0.12\textheight}{%
\begin{tikzpicture}[scale=0.5]

  \begin{scope}
    \useasboundingbox (0,0) rectangle (8,8);
    \draw[step=4cm,thick, scale=1] (0, 0) grid (8, 8);
    \draw[step=1cm,thick, scale=1,xshift=2cm] (0, 0) grid (2, 2);
    \draw[step=2cm,thick, scale=1] (0, 0) grid (4, 4);
    \draw[step=1cm,thick, scale=1,yshift=2cm,xshift=6cm] (0, 0) grid (2, 2);
    \draw[step=2cm,thick, scale=1,yshift=4cm,xshift=4cm] (0, 0) grid (4, 4);
    \draw[step=1cm,thick, scale=1,yshift=6cm,xshift=4cm] (0, 0) grid (2, 2);
    \draw[step=1cm,thick, scale=1,yshift=6cm,xshift=6cm] (0, 0) grid (2, 2);
    \draw[step=1cm,thick, scale=1,yshift=4cm,xshift=6cm] (0, 0) grid (2, 2);
    \draw[step=1cm,thick, scale=1,yshift=2cm,xshift=6cm] (0, 0) grid (2, 2);
    \draw[step=1cm,thick, scale=1,yshift=0cm,xshift=6cm] (0, 0) grid (2, 2);
    \draw[step=1cm,thick, scale=1,yshift=2cm,xshift=4cm] (0, 0) grid (2, 2);
    \draw[step=1cm,thick, scale=1,yshift=0cm,xshift=4cm] (0, 0) grid (2, 2);
\node at (2,6) {$\bullet$};
\node[fill=darkgray,minimum size=0.47cm] at (3.5,1.5) {};
\node[fill=darkgray,minimum size=0.47cm] at (3.5,0.5) {};
\node[fill=darkgray,minimum size=0.47cm] at (4.5,6.5) {};
\node[fill=darkgray,minimum size=0.47cm] at (4.5,7.5) {};
\node[fill=darkgray,minimum size=0.47cm] at (6.5,2.5) {};
\node[fill=darkgray,minimum size=0.47cm] at (6.5,3.5) {};
\node[fill=darkgray,minimum size=0.47cm] at (6.5,4.5) {};
\node[fill=darkgray,minimum size=0.47cm] at (7.5,4.5) {};
\node[fill=darkgray,minimum size=0.47cm] at (7.5,3.5) {};
\node[fill=darkgray,minimum size=0.47cm] at (7.5,2.5) {};
\node[fill=darkgray,minimum size=0.47cm] at (5.5,3.5) {};
\node[fill=darkgray,minimum size=0.47cm] at (5.5,2.5) {};
  \end{scope}
    
\end{tikzpicture}}
\caption{\#10 - mDASPAR}
\label{fig:visitinst10-3}
\end{subfigure}
\end{figure}
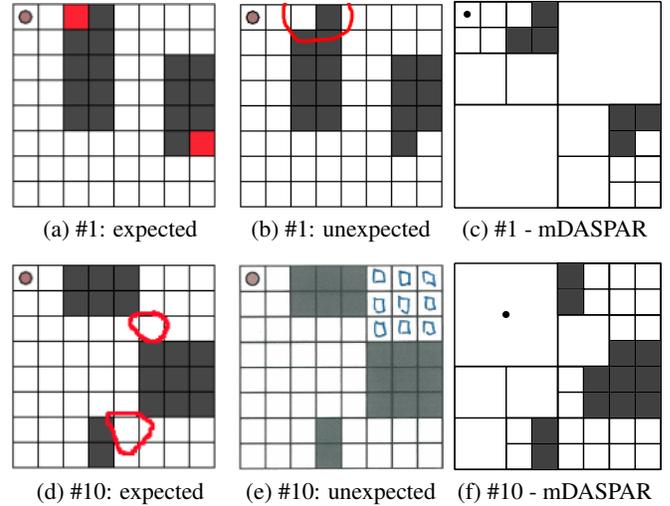

\leanparagraph{Visitall}
Most of the users pick two dead-end cells in the instances (if such occur) as an explanation for unsatisfiability. However, the explanations are given by marking these dead-end cells, instead of the obstacles surrounding them (e.g., Fig.~\ref{fig:visitinst1-1}), which are the actual cause for them to be dead-end cells. Even with abstraction mappings that also distinguish the surrounding obstacles, the corresponding abstract program is still satisfiable. mDASPAR needs to refine further to distinguish the neighboring cells (as in Fig.~\ref{fig:visitinst1-3}), to realize that it can only pass through one grid-cell when reaching the dead-end cells, and thus achieve unsatisfiability. 

Some instances do not contain two dead-end cells, but single-cell passages to some regions. Fig.~\ref{fig:visitinst10-1} shows an entry that distinguishes these passages, while again focusing only on the cells themselves. 
For these instances, the results of mDASPAR are quite different. 
For V$_{\textup{KT}}$, even the abstraction mapping that extends the explanation in Fig.~\ref{fig:visitinst10-1} by distinguishing the obstacles can not achieve unsatisfiability.
This is due to guessing a set of $\mi{move}$ atoms, which achieves that every cell is visited, but actually does not have a corresponding original order of movements.
The abstraction does not achieve unsatisfiability for V as well. The abstract encoding is able to compute a plan that traverses over different-sized regions by avoiding the constraints due to uncertainty.
Fig.~\ref{fig:visitinst10-3} shows the best abstraction achieved for V$_{\textup{KT}}$. It distinguishes all the cells in the one-passage-entry regions to realize that a desired action sequence can not be found.

\paragraph{Observations} 
The abstraction method can demonstrate the capability of human-like focus on certain parts of the grid to show the unsolvability reason. However, humans are also implicitly making use of their background knowledge and do not need to explicitly state the relations among the objects. Empowering the machine with such capabilities remains a challenge. The study also showed the difference in understanding the meaning of ``explanation". For some, showing the solution to get rid of unsolvability is also seen as an explanation. This difference in understanding shows that one needs to clearly specify what they want (e.g., ``mark only the obstacles that cause to have unreachable cells"), 
which would then deviate from studying the meaning of explanation.

\section{Conclusion}

In this paper we introduced a novel approach for using abstraction to focus on certain aspects of problem solving in ASP. The method allows for making use of the structure in the problem for describing the abstraction, and shows an automated way of starting with an initial abstraction and achieving an abstraction with a concrete answer. 
We use this approach to obtain (quad-tree style) abstractions that zoom in to the reason for unsolvability of grid-cell problem instances. The user study shows that these abstractions match the intuition behind human explanations, and can be used to obtain explanations for unsolvability. Although, achieving the various levels of abstraction in human explanations is a challenge.

\emleanparagraph{Related Work} The most relevant work of abstraction
in AI can be seen in the planning community, 
which mostly focuses on abstracting the state space
\cite{sacerdoti1974planning,helmert2014merge,McIlraith19}. A syntactic
approach with a hierarchical view of abstraction over the domain has
not been considered. The recent work on
explanations for unsolvability \cite{sreedharan2019couldn} focuses on projecting out irrelevant objects, similar to \cite{zgskr18}.

\bibliographystyle{named}

\begin{thebibliography}{}

\bibitem[\protect\citeauthoryear{Brain \bgroup \em et al.\egroup
  }{2007}]{brain2007debugging}
Martin Brain, Martin Gebser, J{\"o}rg P{\"u}hrer, Torsten Schaub, Hans Tompits,
  and Stefan Woltran.
\newblock Debugging ASP programs by means of {ASP}.
\newblock In {\em Proc. LPNMR}, pp.\  31--43. Springer, 2007.

\bibitem[\protect\citeauthoryear{Brewka \bgroup \em et al.\egroup
  }{2011}]{aspglance11}
Gerhard Brewka, Thomas Eiter, and Mirosław Truszczyński.
\newblock Answer set programming at a glance.
\newblock {\em Comm. ACM}, 54(12):92--103, 2011.

\bibitem[\protect\citeauthoryear{Cabalar \bgroup \em et al.\egroup
  }{2014}]{cabalar2014causal}
Pedro Cabalar, Jorge Fandinno, and Michael Fink.
\newblock Causal graph justifications of logic programs.
\newblock {\em TPLP}, 14(4-5):603--618, 2014.

\bibitem[\protect\citeauthoryear{Clarke \bgroup \em et al.\egroup
  }{2003}]{clarke03}
Edmund Clarke, Orna Grumberg, Somesh Jha, Yuan Lu, and Helmut Veith.
\newblock Counter\-exam\-ple-guided abstraction refinement for symbolic model
  checking.
\newblock {\em J.\ ACM}, 50(5):752--794, 2003.

\bibitem[\protect\citeauthoryear{Craik}{1952}]{craik1952nature}
Kenneth James~Williams Craik.
\newblock {\em The nature of explanation}, volume 445.
\newblock CUP Archive, 1952.

\bibitem[\protect\citeauthoryear{Erdem \bgroup \em et al.\egroup
  }{2016}]{DBLP:journals/aim/ErdemGL16}
Esra Erdem, Michael Gelfond, and Nicola Leone.
\newblock Applications of answer set programming.
\newblock {\em {AI} Magazine}, 37(3):53--68, 2016.

\bibitem[\protect\citeauthoryear{Faber \bgroup \em et al.\egroup
  }{2004}]{FLP04}
Wolfgang Faber, Nicola Leone, and Gerald Pfeifer.
\newblock Recursive aggregates in disjunctive logic programs: Semantics and
  complexity.
\newblock In {\em Proc. JELIA}, pp.\  200--212. Springer, 2004.

\bibitem[\protect\citeauthoryear{Gebser \bgroup \em et al.\egroup
  }{2008}]{gebser2008meta}
Martin Gebser, J{\"o}rg P{\"u}hrer, Torsten Schaub, and Hans Tompits.
\newblock A meta-programming technique for debugging answer-set programs.
\newblock In {\em AAAI}, 
pp.\  448--453, 2008.

\bibitem[\protect\citeauthoryear{Gelfond and Lifschitz}{1988}]{gelf-lifs-88}
M.~Gelfond and V.~Lifschitz.
\newblock The stable model semantics for logic programming.
\newblock In {\em ICLP/SLP}, pp.\  1070--1080, 1988.

\bibitem[\protect\citeauthoryear{Giunchiglia and
  Walsh}{1992}]{giunchiglia1992theory}
Fausto Giunchiglia and Toby Walsh.
\newblock A theory of abstraction.
\newblock {\em Artificial Intelligence}, 57(2-3):323--389, 1992.

\bibitem[\protect\citeauthoryear{Helmert \bgroup \em et al.\egroup
  }{2014}]{helmert2014merge}
Malte Helmert, Patrik Haslum, J{\"o}rg Hoffmann, and Raz Nissim.
\newblock Merge-and-shrink abstraction: A method for generating lower bounds in
  factored state spaces.
\newblock {\em J.\ ACM}, 61(3), 2014.

\bibitem[\protect\citeauthoryear{Illanes and McIlraith}{2019}]{McIlraith19}
León Illanes and Sheila~A. McIlraith.
\newblock Generalized planning via abstraction: Arbitrary numbers of objects.
\newblock In {\em AAAI}, 2019.

\bibitem[\protect\citeauthoryear{Johnson-Laird}{1983}]{johnson1983mental}
Philip~Nicholas Johnson-Laird.
\newblock {\em Mental models: Towards a cognitive science of language,
  inference, and consciousness}.
\newblock
Harvard Univ.\ Press, 1983.

\bibitem[\protect\citeauthoryear{Kambhampati and
  Davis}{1986}]{kambhampati1986multiresolution}
Subbarao Kambhampati and Larry Davis.
\newblock Multiresolution path planning for mobile robots.
\newblock {\em IEEE Journal on Robotics and Automation}, 2(3):135--145, 1986.

\bibitem[\protect\citeauthoryear{Knoblock}{1990}]{knoblock1990learning}
Craig~A Knoblock.
\newblock Learning abstraction hierarchies for problem solving.
\newblock In {\em AAAI}, pp.\  923--928, 1990.

\bibitem[\protect\citeauthoryear{Oetsch \bgroup \em et al.\egroup
  }{2010}]{oetsch2010catching}
Johannes Oetsch, J{\"o}rg P{\"u}hrer, and Hans Tompits.
\newblock Catching the ouroboros: On debugging non-ground answer-set programs.
\newblock {\em TPLP}, 10(4-6):513--529, 2010.

\bibitem[\protect\citeauthoryear{Pontelli \bgroup \em et al.\egroup
  }{2009}]{pontelli2009justifications}
Enrico Pontelli, Tran~Cao Son, and Omar Elkhatib.
\newblock Justifications for logic programs under answer set semantics.
\newblock {\em TPLP}, 9(1):1--56, 2009.

\bibitem[\protect\citeauthoryear{Sacerdoti}{1974}]{sacerdoti1974planning}
Earl~D Sacerdoti.
\newblock Planning in a hierarchy of abstraction spaces.
\newblock {\em Artificial Intelligence}, 5(2):115--135, 1974.

\bibitem[\protect\citeauthoryear{Saitta and
  Zucker}{2013}]{saitta2013abstraction}
Lorenza Saitta and Jean-Daniel Zucker.
\newblock {\em Abstraction in artificial intelligence and complex systems},
  volume 456.
\newblock Springer, 2013.

\bibitem[\protect\citeauthoryear{Saribatur and Eiter}{2018}]{zgskr18}
Zeynep~G. Saribatur and Thomas Eiter.
\newblock Omission-based abstraction for answer set programs.
\newblock In {\em Proc. KR}, pp.\  42--51, 2018.

\bibitem[\protect\citeauthoryear{Saribatur \bgroup \em et al.\egroup
  }{2019}]{zgs19jelia}
Zeynep~G. Saribatur, Peter Sch{\"{u}}ller, and Thomas Eiter.
\newblock Abstraction for non-ground answer set programs.
\newblock In {\em Proc. JELIA}, 
pp.\  576--592. Springer, 2019.

\bibitem[\protect\citeauthoryear{Schulz and Toni}{2013}]{schulz2013aba}
Claudia Schulz and Francesca Toni.
\newblock {ABA}-based answer set justification.
\newblock {\em TPLP}, 13(4-5-Online-Supplement), 2013.

\bibitem[\protect\citeauthoryear{Sreedharan \bgroup \em et al.\egroup
  }{2019}]{sreedharan2019couldn}
Sarath Sreedharan, Siddharth Srivastava, David Smith, and Subbarao Kambhampati.
\newblock Why Can’t You Do That HAL? Explaining Unsolvability of Planning Tasks.
\newblock In {\em Proc. IJCAI}, 2019.

\end{thebibliography}

\appendix

\section{Computing Joint Abstract Relation Types}
\label{app:joint}


Abstract relations can be easily employed with abstraction mappings over
several sorts in the domain as $m : D_1 \times \dots \times D_n
\rightarrow \hat{D}_1 \times \dots \times \hat{D}_n$. If
a rule has relations over the sorts, a joint abstract
relation combining them must be computed. 

Assuming for simplicity a uniform arity $k$,
the abstract $k$-tuple
relations
are computed 
by
{\small
\def\mydots{...\,}
\begin{align}
\widehat{\rel}_i((\hat{d}_1^1,\mydots,&\hat{d}_1^n),\mydots,(\hat{d}_k^1,\mydots,\hat{d}_k^n))
\leftarrow \rel_i(d_1^i,\mydots, d_k^i),\nonumber\\ 
&\qquad{\textstyle\bigwedge_{j=1}^k} m(((d_j^1,\mydots,d_j^n)),((\hat{d}_j^1,\mydots,\hat{d}_j^n))).\nonumber
\end{align}
}
\noindent for $i=1,\ldots, n$. 
We compute the types of these auxiliary abstract relations, for objects $\hat{c}_j = (\hat{d}_j^1,\dots,\hat{d}_j^n), 1\leq j \leq k$.
{\small
\begin{align}
\tau_{\textup{I}}^{\widehat{\rel}_i}(\hat{c}_1,\dots,\hat{c}_k) &\leftarrow \widehat{\rel}_i(\hat{c}_1,\dots,\hat{c}_t), \mi{not}\ \neg\widehat{\rel}_i(\hat{c}_1,\dots,\hat{c}_t)\nonumber\\
\tau_{\textup{II}}^{\widehat{\rel}_i}(\hat{c}_1,\dots,\hat{c}_t) &\leftarrow \mi{not}\ \widehat{\rel}_i(\hat{c}_1,\dots,\hat{c}_t), \neg\widehat{\rel}_i(\hat{c}_1,\dots,\hat{c}_t)\nonumber\\
\tau_{\textup{III}}^{\widehat{\rel}_i}(\hat{c}_1,\dots,\hat{c}_t) &\leftarrow \widehat{\rel}_i(\hat{c}_1,\dots,\hat{c}_t), \neg\widehat{\rel}_i(\hat{c}_1,\dots,\hat{c}_t)\nonumber
\end{align}
}

The types of the joint abstract relation $\widehat{\rel}$ over the
objects $\hat{c}_j = (\hat{d}_j^1,\dots,\hat{d}_j^n), 1\leq j \leq k$ (i.e.\ $\tau_{\textup{I}}^{\widehat{rel}}$ and $\tau_{\textup{III}}^{\widehat{rel}}$),
are then computed as below.
{\small
\begin{align}
\tau_{\textup{I}}^{\widehat{rel}}(\hat{c}_1,\dots,\hat{c}_k) &\leftarrow \tau_{\textup{I}}^{\widehat{\rel}_1}(\hat{c}_1,\dots,\hat{c}_k),\dots, \tau_{\textup{I}}^{\widehat{\rel}_n}(\hat{c}_1,\dots,\hat{c}_k)\nonumber\\
\tau_{\textup{III}}^{\widehat{rel}}(\hat{c}_1,\dots,\hat{c}_k)
&\leftarrow
\tau_{\textup{III}}^{\widehat{\rel}_i}(\hat{c}_1,\dots,\hat{c}_k),\nonumber\\
& \quad{\textstyle\bigwedge_{j=1: j\neq i}^n}\mi{not}\ \tau_{\textup{II}}^{\widehat{\rel}_j}(\hat{c}_1,\dots,\hat{c}_k), 1\leq i \leq n.\nonumber
\end{align}
}

Note that for the joint abstract relation $\widehat{rel}$, type $\tau_{\textup{II}}^{\widehat{rel}}$ computation is not needed, as the abstract rule construction only deals with types I and III.

\end{document}